\documentclass[10pt,twocolumn]{article}
\usepackage[margin=0.72in,top=0.78in,bottom=0.78in]{geometry}
\usepackage{amsmath,amssymb,booktabs,array,microtype,float,needspace}
\usepackage[T1]{fontenc}
\usepackage{times}
\usepackage[hidelinks]{hyperref}
\title{DynaDFlash: Dynamic Feature Projection for Block Diffusion Speculators}
\author{Aadi Kulshrestha\\\texttt{a7kulshrestha@uwaterloo.ca} \and Reese Chong\\\texttt{r25chong@uwaterloo.ca}}
\date{November 12, 2026}

\begin{document}
\maketitle
\vspace{-1.8em}

\begin{abstract}
DFlash trains a parallel draft model on hidden features from its target model. Inspired by the previous work of DFlare that introduced per layer combinations of target hidden states, we test whether context dependent weighting of these states helps improve acceptance length of block diffusion speculators. We trained both a DFlash baseline for the Qwen 3 of block size 16, and a dynamic model that computes gates for each layer to affect the magnitude of hidden features before projection. While mean acceptance length between the dynamic model and the DFlash baseline didn't significantly differ, we notice that the dynamic model excels for drafting in the context of math. In an evaluation of 700 prompts from the MetaMathQA and Orca-Math datasets, the mean accepted length improves by ~5\%. We also found this lead to a 4.1\% higher throughput for a subset of the same dataset in the context of speculative decoding.
\end{abstract}

\section{Introduction}

Speculative decoding uses a smaller draft model to predict tokens that are then verified by a larger target model~\cite{leviathan2023specdec,chen2023specsample}. DFlash does this by training a block diffusion model to draft multiple tokens in parallel, using hidden states from the target model as additional context that is injected at each layer~\cite{chen2026dflash}.

DFlare builds on this idea by learning different combinations of target hidden states for different draft layers~\cite{zhang2026dflare}. However, these combinations remain fixed for each layer after training. We wanted to test whether allowing the weights to change depending on the input context could further improve drafting.

We modify DFlash by adding a small gate that computes weights for the target hidden states before they are projected. We then train both the original DFlash architecture and our modified version under the same conditions and compare their acceptance lengths and decoding throughput.

\section{Background}

Autoregressive LLMs generate tokens one at a time. Each token requires a new forward pass through the target model, which slows down generation. Speculative decoding allows us to produce more tokens for each forward pass of the target model. A smaller model first drafts several tokens, and the target model checks them together in a single forward pass~\cite{leviathan2023specdec,chen2023specsample}.

Under greedy decoding, a draft token is accepted if it matches the highest likelihood token the target would have produced. Tokens are accepted until the first mismatch, at which point the target produces the next token instead. The final output is therefore the same as ordinary greedy decoding. For sampling, speculative decoding uses a different acceptance formula to preserve the target model's output distribution~\cite{leviathan2023specdec,chen2023specsample}.

The speedup comes from verifying several tokens in parallel rather than running the target separately for every token. A useful approximation for average time per generated token is~\cite{chen2026dflash}
\begin{equation}
L \approx \frac{T_{\mathrm{draft}}+T_{\mathrm{verify}}}{\tau},
\label{eq:latency}
\end{equation}
where $T_{\mathrm{draft}}$ is the time it takes for the drafter to propose a block, $T_{\mathrm{verify}}$ is the time to check it via the target model, and $\tau$ is the average number of tokens produced per verification cycle (the accepted tokens + 1). Examining this approximation it becomes clear that to improve throughput we must balance the latency of the drafter with its acceptance length.

DFlash reduces drafting time by predicting the next 15 tokens in parallel from a block of masked positions, rather than generating each draft token one at a time~\cite{chen2026dflash}. The target then verifies the block. This makes both drafting time and acceptance length important when comparing draft models.

\section{Models}

DFlash takes hidden states from layers 2, 10, 18, 26, and 34 of the target model, concatenates them, and projects them into the draft model.

\Needspace{11\baselineskip}
Let $\mathbf h_t^{(l)}$ be the hidden state from target layer $l$ at token position $t$. DFlash combines these states using a learned projection $W_c$:
\begin{equation}
\begin{aligned}
\mathbf z_t &= [\mathbf h_t^{(2)};\mathbf h_t^{(10)};\mathbf h_t^{(18)};\\
&\hspace{2.1em}\mathbf h_t^{(26)};\mathbf h_t^{(34)}],\\
\mathbf c_t &= \mathrm{RMSNorm}(W_c\mathbf z_t).
\end{aligned}
\label{eq:dflash}
\end{equation}
The resulting context feature $\mathbf c_t$ is passed to every draft layer through its key and value (KV) cache. The same projection is used for every token and every draft layer~\cite{chen2026dflash}.

DFlare instead learns a different linear combination of target states for each draft layer $i$~\cite{zhang2026dflare}:
\begin{equation}
\begin{aligned}
\boldsymbol\alpha^{(i)} &= \mathrm{softmax}(\mathbf w^{(i)}),\\
\mathbf f_t^{(i)} &= \mathrm{RMSNorm}\!\left(\sum_j \alpha_j^{(i)}\mathbf h_t^{(j)}\right).
\end{aligned}
\label{eq:dflare}
\end{equation}
The weights depend on the draft layer $i$, but are fixed across token positions $t$ once the model is trained. Our experiment instead keeps DFlash's projection and changes the weights based on the current hidden state.

Our modification keeps the same projection as DFlash but adds a linear gate before it. The gate takes the hidden state from layer 34 and computes five weights, one for each of the selected target layers. Each hidden state is multiplied by its corresponding weight before concatenation and projection.

We write the gate and the resulting projection as
\begin{equation}
\begin{aligned}
\mathbf g_t &= G_\theta(\mathbf h_t^{(34)}),\\
\widetilde{\mathbf z}_t
  &= [g_{t,1}\mathbf h_t^{(2)};\ldots;
      g_{t,5}\mathbf h_t^{(34)}],\\
\widetilde{\mathbf c}_t
  &= \mathrm{RMSNorm}(W_c\widetilde{\mathbf z}_t).
\end{aligned}
\label{eq:dynamic}
\end{equation}
$G_\theta$ is the learned gate and $g_{t,j}$ is the weight assigned to the $j$-th selected target layer at position $t$. The projection $W_c$ is still learned in the same way as DFlash, only its input magnitude is adjusted. The gate is initialized so that $\mathbf g_t=\mathbf 1$, making Equation~\ref{eq:dynamic} identical to Equation~\ref{eq:dflash} at the start.

Unlike the fixed projection in DFlash, these weights are computed separately for each token, allowing the contribution of each target layer to depend on the context.

The gate introduces only 20,480 additional parameters.

\section{Evaluation}

\textbf{Training.} We train both models using a frozen Qwen3-8B target and the same 792,000 training examples in the same order. Both use DFlash's original training methodology with a block size of 16.

Training runs for one epoch, consisting of 24,750 updates with a global batch size of 32 on two H100 GPUs. We use 990 warmup steps, a peak learning rate of $6\times10^{-4}$, and cosine learning rate decay. We evaluate the final checkpoints after checking their training histories.

\textbf{Math evaluation.} After observing stronger performance on math prompts, we conduct a separate test using 700 held-out prompts from MetaMathQA and Orca-Math~\cite{yu2024metamath,mitra2024orcamath}. Of these, 647 come from MetaMathQA and 53 from Orca-Math. 

For each prompt, we provide the question without its answer. The target model generates the first token, and the draft model predicts the next 15 tokens. We then measure how many tokens match the target model's greedy predictions, including the first token generated by the target. We count acceptances from the beginning of the draft and stop at the first mismatch.

We use mean accepted length as our main metric to measure how closely the draft model predict the target's most likely token.

For our block size of 16, we count the first target token plus the number of draft tokens accepted before the first mismatch:
\begin{equation}
A = 1 + \max\bigl\{k: \hat y_j=y_j^*\ 
\text{for all }1\leq j\leq k\bigr\},
\label{eq:acceptance}
\end{equation}
where $0\leq k\leq15$, $\hat y_j$ is a draft token, and $y_j^*$ is the target's greedy token at that position. This definition matches the acceptance length reported in our evaluation.

\textbf{Serving.} We also measure decoding throughput using SGLang 0.5.18 on one A100 GPU. We select 24 fixed MetaMathQA prompts and generate 64 tokens per request using greedy decoding.

We test both models with eight concurrent requests. Each experiment is repeated three times using new servers, alternating the order in which the models are tested. Throughput is measured in output tokens per second.

\section{Results}

\begin{table}[H]
\centering
\caption{Mean accepted tokens on 700 math prompts.}
\label{tab:accept}
\small
\begin{tabular}{@{}lrr@{}}
\toprule
Model & Accepted tokens & Change \\
\midrule
DFlash & 2.98 & -- \\
DynaDFlash & \textbf{3.13} & \textbf{+5.0\%} \\
\bottomrule
\end{tabular}
\end{table}

On the 700 math prompts, the dynamic model achieves a mean accepted length of 3.13 tokens, compared to 2.98 for the original DFlash model. This is an improvement of about 5\%, or 0.15 additional tokens per prompt.

The 95\% confidence interval for the difference is 0.07 to 0.22 tokens, with $p<0.001$.

Most of this improvement comes from the 647 MetaMathQA prompts, where the dynamic model accepts an additional 0.16 tokens on average. We do not observe an improvement on the smaller set of 53 Orca-Math prompts.

The results are different on our full evaluation, where the original DFlash model and the dynamic model have no statistically significant difference in acceptance length. This suggests that the benefits of dynamic weighting may depend on the type of input rather than improving drafting across all tasks.

We also observe a throughput improvement when serving eight requests in parallel. The dynamic model generates an average of 1004.5 output tokens per second compared to 965.0 for DFlash, an improvement of 4.1\%.

\begin{table}[H]
\centering
\caption{Throughput is measured in output tokens per second.}
\label{tab:throughput}
\small
\begin{tabular}{@{}lrr@{}}
\toprule
Model & Output tokens/s & Change \\
\midrule
Autoregressive Qwen3-8B & 655.4 & -- \\
DFlash & 965.0 & -- \\
DynaDFlash & 1004.5 & +4.1\% \\
\bottomrule
\end{tabular}
\end{table}

\section{Discussion}

Our results show that dynamic context projection can improve DFlash's acceptance length on math prompts, but the improvement does not extend to all types of prefixes. In particular, the dynamic model performs better on MetaMathQA while not significantly differing in performance from the baseline for the full evaluation.

We also looked at the weights produced by the gate to understand whether certain target layers were more useful for math. On the 700 math prompts, layers 2 had the largest mean gate, and layer 26 had the second largest. However, due to differences in the magnitude of the hidden features for these layers, layer 26 contributed the most to the projection, followed layer 34.

A key limitation is that both models were trained independently and do not match the acceptance length of the published DFlash model by z-lab. This makes the observed improvement in acceptance length less reliable as both models fall below the acceptance length of other DFlash drafters.

\section{Conclusion}

We tested whether dynamically weighting target hidden states before projection could improve DFlash's draft model. Our modified model achieves a 5\% higher mean accepted length on 700 math prompts and a 4.1\% throughput improvement in our serving test.

However, these improvements are not consistent across datasets, and we could not establish a clear relationship between dynamic context to the drafter and model performance. Further experimentation may be required with different methods to inject dynamic information from the target model to the drafter to demonstrate a gain across various contexts.

\section*{Acknowledgements}

We thank Modal for supporting our work through a \$5,000 research grant. All model training and evaluation was conducted on Modal's platform.

\appendix
\section{Reproducibility and ablation}

The training and evaluation code is in the \href{https://github.com/midas-llm-research/DynaDFlash}{DynaDFlash repository}. The trained \href{https://huggingface.co/midas-llm-research/qwen3-8b-dflash-baseline}{DFlash baseline} and \href{https://huggingface.co/midas-llm-research/qwen3-8b-dynadflash}{DynaDFlash} models are on Hugging Face.

We trained a compact variant that sums the five gated target states before projection. This also makes the projection smaller. On the 700 math prompts, it accepts 2.96 tokens on average, compared with 2.98 for the DFlash baseline and 3.13 for DynaDFlash.

\end{document}